# Performance Analysis of Unsupervised Feature Selection Methods

A. Nisthana Parveen[1]
Research Scholar

H. Hannah Inbarani[2]
Assistant Professor

E.N. Sathish Kumar[3]
Research Scholar

Department of Computer Science
Periyar University, Salem-11.
[1]E-Mail: nisthana@gmail.com
[2]E-Mail: hhinba@gmail.com
[3]E-Mail: en.sathishkumar@yahoo.co.in

*Abstract*

**Feature selection (FS) is a process which attempts to select more informative features. In some cases, too many redundant or irrelevant features may overpower main features for classification. Feature selection can remedy this problem and therefore improve the prediction accuracy and reduce the computational overhead of classification algorithms. The main aim of feature selection is to determine a minimal feature subset from a problem domain while retaining a suitably high accuracy in representing the original features. In this paper, Principal Component Analysis (PCA), Rough PCA, Unsupervised Quick Reduct (USQR) algorithm and Empirical Distribution Ranking (EDR) approaches are applied to discover discriminative features that will be the most adequate ones for classification. Efficiency of the approaches is evaluated using standard classification metrics.**

*Keywords: Feature Selection, Principal Component Analysis, Rough-PCA, Empirical Distribution, Unsupervised Quick Reduct.*

## I. INTRODUCTION

Feature selection, is a problem closely related to dimension reduction. The objective of feature selection is to identify features in the data-set as important, and discard any other feature as irrelevant and redundant information. Since feature selection reduces the dimensionality of the data, it holds out the possibility of more effective & rapid operation of data mining algorithm (i.e. Data Mining algorithms can be operated faster and more effectively by using feature selection).

Conventional supervised FS methods evaluate various feature subsets using an evaluation function or metric to select only those features which are related to the decision classes of the data under consideration. However, for many data mining applications, decision class labels are often unknown or incomplete, thus indicating the significance of unsupervised feature selection. In unsupervised learning, decision class labels are not provided.

Principal Components Analysis (PCA) is the predominant linear dimensionality reduction technique, and it has been widely applied on datasets in all scientific domains. In words, PCA seeks to map or embed data points from a high dimensional space to a low dimensional space while keeping all the relevant linear structure intact. To improve the efficiency and accuracy of data mining task on high dimensional data, the data must be preprocessed by an efficient dimensionality reduction method. Principal Component Analysis (PCA) is a popular linear feature extractor used for unsupervised feature selection based on eigenvectors analysis to identify critical original features for principal component. PCA is a statistical technique for determining key variables in a high dimensional data set that explain the differences in the observations and can be used to simplify the analysis and visualization of high dimensional data set, without much loss of information. Rough set theory is employed to generate reducts, which represent the minimal sets of non-redundant features capable of discerning between all objects, in a multiobjective framework. Rough-PCA approach is the combination of PCA and Rough set theory.

The rest of the paper is organized as follows: Section 2, briefs about the feature selection algorithm such as PCA, Rough-PCA, Unsupervised Quick Reduct and Empirical Distribution. Section 3 explains briefly about experimental analysis and results. Section 4 presents a conclusion for this paper.

## II. FEATURE SELECTION METHODS

### A. Principal Component Analysis

Principal Component Analysis is an unsupervised Feature Reduction method for projecting high dimensional data into a new lower dimensional representation of the data that describes as much of the variance in the data as possible with minimum reconstruction error. Principal Component Analysis is a quantitatively rigorous method for achieving this simplification. The method generates a new set of variables, called principal components. Each principal component is a linear combination of the original variables. All the principal components are orthogonal to each other, so there is no redundant information. The principal components as a whole form an orthogonal basis for the space of the data. Thus we propose unsupervised feature selection algorithms based on eigenvectors analysis to identify critical original features for principal component [5].

PCs are calculated using the Eigen value decomposition of the data covariance matrix/correlation matrix or singular value decomposition of a data matrix. Usually after mean centering the data for each attribute. Covariance matrix is preferred when the variances of variables are very high compared to correlation. It would be better to choose the type correlation when the variables are of different types. Similarly the SVD method is used for numerical accuracy.

Singular value decomposition (SVD) can be looked at from three mutually compatible points of view. On the one hand, we can see it as a method for transforming correlated variables into a set of uncorrelated ones that better expose the various relationships among the original data items. At the same time, SVD is a method for identifying and ordering the dimensions along which data points exhibit the most variation.

SVD and PCA are common techniques for analysis of multivariate data, and gene expression data are well suited to analysis using SVD/PCA. We can use SVD to perform PCA. SVD is based on a theorem from linear algebra which says that a rectangular matrix X can be broken down into the product of three matrices – an orthogonal matrix U, a diagonal matrix S, and the transpose of an orthogonal matrix S, and the transpose of an orthogonal matrix V. The theorem is usually presented something like this:

$$A_{mm} = U_{mm} S_{mn} V_{nn}^T \quad (1)$$

where $U^T U = 1$, $V^T V = 1$; the columns of U are of U are orthonormal eigenvectors of $AA^T$, the columns of V are orthonormal eigenvectors of $A^T A$, and S is a diagonal matrix containing the square roots of eigen values from U or V in descending order. The resulting algorithm is given below.

**Algorithm: PCA**
**Input: Data Matrix**
**Output: Reduced set of features**
**Step-1**: X ← Create N x d data matrix, with one row vector $x_n$ per data point.
**Step-2**: X subtract mean $x$ from each row vector $x_n$ in X.
**Step-3**: Σ ← covariance matrix of X.
**Step-4**: Find eigenvectors and eigen values of Σ.
**Step-5**: PC's ← the M eigenvectors with largest eigen values.
**Step-6**: Output PCs.

Algorithm1: Principal Component Analysis

*B. Rough-PCA*

1. Rough Set Theory

Rough set theory (RST) has been used as a tool to discover data dependencies and to reduce the number of attributes contained in a dataset using the data alone, requiring no additional information [3] [4]. Over the past ten years, RST has become a topic of great interest to researchers and has been applied to many domains. Given a dataset with discretized attribute values, it is possible to find a subset (termed a reduct) of the original attributes using RST that are the most informative; all other attributes can be removed from the dataset with minimal information loss. An information table is defined as a tuple T = (U, A) where U and A are two finite, non-empty sets, U the universe of primitive objects and A the set of attributes. Each attribute or feature a∈ A is associated with a set Va of its value, called the domain of a. We may partition the attribute set A into two subsets C and D, called condition and decision attributes, respectively[8].

Let P ⊂ A be a subset of attributes. The indiscernibility relation, denoted by IND (P), is an equivalence relation defined as:

IND (P) = {(x, y) ∈ U×U: ∀ a∈ P, a(x) = a(y)
$$\quad (2)$$

where a(x) denotes the value of feature a of object x. If (x, y) ∈ IND (P), x and y are said to be indiscernible with respect to P.

The family of all equivalence classes of IND (P) (Partition of U determined by P) is denoted by U/IND (P). Each element in U/IND (P) is a set of indiscernible objects with respect to P. Equivalence classes U/IND(C) and U/IND (D) are called

condition and decision classes, and it can be calculated as follows:

$$U/IND(P) = \otimes \{a \in P: U/IND(\{a\})\} \qquad (3)$$

Where
$$A \otimes B = \{X \cap Y: \forall X \in A, \forall Y \in B, X \cap Y \neq \emptyset\} \qquad (4)$$

If $(x, y) \in IND(P)$, then $x$ and $y$ are indiscernible by attributes from $P$. The equivalence classes of the $P$-indiscernibility relation are denoted $[x] P$.

A rough set is defined by the lower and upper approximations of a concept. The lower approximation contains all elements that necessarily belong to the concept, while the upper approximation contains those that possibly belong to the concept. In rough set theory, a concept is considered a classical set.

Let $X \subseteq U$. $X$ can be approximated using only the information contained within $P$ by constructing the P-*lower* and P-*upper* approximations of $X$:

$$\underline{P}X = \{x \mid [x]p \subseteq X\} \qquad (5)$$

$$\overline{P}X = \{x \mid [x]p \cap X \neq \emptyset\} \qquad (6)$$

Where $[x]p$ denotes the equivalence class of object $x \in U$ relative to $I_p$, are called the P-lower and P-upper approximation of X and denoted by $\underline{P}X$, $\overline{P}X$ respectively.

Let P and Q be equivalence relations over U, then the positive, negative and boundary regions can be defined as:

$$POSp(Q) = U_{X \in U/Q} \underline{P}X \qquad (7)$$

The positive region contains all objects of U that can be classified to classes of U/Q using the information in attributes P.

Rough set reducts can be found by using degree of dependency or by using discernibility matrix.

$$k = \gamma P(Q) = \frac{|POSp(Q)|}{|U|} \qquad (8)$$
Where
$$POSp(Q) = U_{X \in U/Q} \underline{P}X \qquad (9)$$

The reduction of attributes is achieved by comparing equivalence relations generated by sets of attributes. Attributes are removed so that the reduced set provides the same predictive capability of the decision feature as the original. A reduct is defined as a subset of minimal cardinality $Rmin$ of the conditional attribute set C such that $\gamma R(D) = \gamma C(D)$.

$$R = \{X: X \subseteq C, \gamma X(D) = \gamma C(D)\} \qquad (10)$$

$$Rmin = \{X: X \in R, \forall Y \in R, |X| \leq |Y|\} \qquad (11)$$

2. Rough-PCA Algorithm

Principal component analysis is an unsupervised linear feature reduction method for projecting high-dimensional data into a low-dimensional space with minimum loss of information. It discovers the directions of maximal variances in the data. The Rough set approach to feature selection is used to discover the data dependencies and reduction in the number of attributes contained in a dataset using the data alone, requiring no additional information. For selecting discriminative features from principal components, the Rough set theory can be applied jointly with PCA, which guarantees that the selected principal components will be the most adequate for classification. We call this method Rough-PCA. The method is successfully applied for choosing the principal features and then applying the upper and lower approximations to find the reduced set of features. The resulting algorithm is given below [5].

**Algorithm: Rough PCA**
**Input: Data Matrix**
**Output: Reduced set of features**
**Step-1**: Normalize the original data set.
**Step-2**: Calculate the Principal Components using Singular Value Decomposition of the Normalized data matrix.
**Step-3**: Determine the number of meaningful PCs to retain.
**Step-4**: Find the reduced data set using the reduced PCs.
**Step-5**: Discretize the data set.
**Step-6**: Find the reduct using Rough set theory (RST).

Algorithm 2: Rough-PCA

*C. Empirical Distribution Ranking*

Let $(x_1, x_2 \ldots x_n)$ be iid or independent identically distributed real random variables with common cdf F (t). Then the empirical distribution function is defined as

$$F_n(t) = \frac{1}{n} \sum_{i=1}^{n} I_{[X_i \leq t]} \qquad (12)$$

Where t is the mean of $X_i$, $I_A$ is the so-called indicator random variable which is defined to be equal to 1 when the property A holds, and equal to 0 otherwise. Thus, while the distribution function gives as a function of t the probability with which each of the random variables $X_i$ will be $\leq t$, the empirical distribution function calculated from data gives the relative frequency with which the observed values are $\leq t$. Sorting the values of $F_n(t)$, then choosing the minimum value attributes for ranking [3].

---

**Algorithm: EDR**
**Input: Data Matrix**
**Output: Reduced set of features**
**Step-1**: Sort the original data set.
$$x'_{i1} < x'_{i2} < \cdots x'_{im}$$
**Step-2**: Calculate the mean value of sorted data
**Step-3**: Find ED using $F_n(t)$.
$$F_n(t) = \frac{1}{n}\sum_{i=1}^{n} I_{[X_i \leq t]}$$
**Step-4**: Rank the features based on ED.

---

*D. Unsupervised Quick Reduct (USQR) Algorithm*

The USQR algorithm attempts to calculate a reduct without exhaustively generating all possible subsets. It starts off with an empty set and adds in turn, one at a time, those attributes that result in the greatest increase in the rough set dependency metric, until this produces its maximum possible value for the dataset [2]. According to the algorithm, the mean dependency of each attribute subset is calculated and the best candidate is chosen:

$$\gamma_P(a) = \frac{|POS_P(a)|}{|U|}, \forall a \in A.$$

---

**Algorithm: USQR (C)**
C, the set of all conditional features;
(1) R ← { }
(2) do
(3) T ← R
(4)     $\forall x \in (C - R)$
(5)     $\forall y \in C$
(6)     $\gamma_{R \cup \{x\}}(y) = \frac{|POS_{R \cup \{x\}}(y)|}{|U|}$
(7)     if $\overline{\gamma_{R \cup \{x\}}(y)}, \forall y \in C > \overline{\gamma_{T}(y)}, \forall y \in C$
(8)         T ← R ∪ {x}
(9) R ← T
(10) until $\overline{\gamma_R(y)}, \forall y \in C = \overline{\gamma_C(y)}, \forall y \in C$
(11) return R

Algorithm 4: Unsupervised QuickReduct

---

## III. EXPERIMENTAL RESULTS

This section presents the results of experimental studies using both crisp-valued and real-valued data sets. Initially we evaluated the algorithm on a datasets, which are available in the UCI machine learning repository. In our experiment, PCA, Rough-PCA, Unsupervised Quick Reduct and Empirical distribution were implemented using **Matlab**. A short experimental evaluation for benchmark datasets is presented. The information of the data sets contains names of dataset, number of objects, number of classes and number of attributes, which are given in Table 1.

TABLE 1: DATASET INFORMATION

| Index | Dataset | Instances | Class | Attr_size |
|---|---|---|---|---|
| 1 | Lung Cancer | 32 | 3 | 56 |
| 2 | Breast Cancer | 569 | 2 | 30 |
| 3 | Diabetes | 768 | 2 | 8 |
| 4 | Heart | 270 | 2 | 13 |
| 5 | Ecoli | 336 | 8 | 7 |

The features are reduced by the PCA, Rough-PCA, Unsupervised Quick Reduct and Empirical distribution algorithms. The selected features are tabulated in table 2.

TABLE 2: SELECTED FEATURES

| Datasets | Reduced attributes obtained by PCA | Reduced attributes obtained by Rough PCA | Reduced attributes obtained by Empirical distribution | Reduced attributes obtained by USQR |
|---|---|---|---|---|
| Lung Cancer | (1 to 17) | 2,3,5,6 | 7,18,39,26,22,33 | 9,13,26,35,36 |
| Breast Cancer | 1,2,3,4,5,6 | 1,2,3,4,5,6 | 5,8,22,9,2 | 8,9 |
| Diabetes | 1,2,3 | 1,3 | 3,4,6,2 | 2,7,8 |
| Heart | 1,2.3,4,5 | 1,3,5 | 2,8,11,1 | 1,5,8 |
| Ecoli | 1,2,3 | 1,2 | 1,5,6,2 | 1,2,6 |

## A. Weka Classification

The Waikato Environment for Knowledge Analysis (Weka) is a comprehensive suite of Java class libraries that implement many state-of-the-art machine learning and data mining algorithms. Weka is freely available on the World-Wide Web and accompanies a new text on data mining [2] which documents and fully explains all the algorithms it contains. Applications written using the Weka class libraries can be run on any computer with a Web browsing capability; this allows users to apply machine learning techniques to their own data regardless of computer platform. Tools are provided for pre-processing data, feeding it into a variety of learning schemes, and analyzing the resulting classifiers and their performance [4].

An important resource for navigating through Weka is its on-line documentation, which is automatically generated from the source. The primary learning methods in Weka are "classifiers", and they induce a rule set or decision tree that models the data. Weka also includes algorithms for learning association rules and clustering data.

The core package contains classes that are accessed from almost every other class in Weka. The most important classes in it are *Attribute*, *Instance*, and *Instances*. An object of class Attribute represents an attribute—it contains the attribute's name, its type, and, in case of a nominal attribute, it's possible values. An object of class Instance contains the attribute values of a particular instance; and an object of class Instances contains an ordered set of instances—in other words, a dataset.

In this paper we have taken the classifiers such as JRip, J48, RBFN, Decision Table, K-Star and Naive Bayes. The determined datasets that are taken from feature selection methods such as Rough PCA, PCA, USQR and Empirical distribution are classified using the above referred classifiers. Table 3, 4, 5, 6 shows the correctly classified instances of mentioned feature selection methods respectively.

TABLE 3: CLASSIFICATION ACCURACY FOR DIABETES

| Classifiers | PCA | Rough-PCA | EDR | USQR |
|---|---|---|---|---|
| JRip | 72.78 | 67.8385 | 76.1719 | 74.2185 |
| J48 | 72.526 | 67.8385 | 73.8281 | 71.4844 |
| RBFN | 74.4792 | 65.625 | 75.9115 | 75.7813 |
| Naive Bayes | 75.1302 | 65.3646 | 76.3021 | 75.651 |
| Decision Table | 73.5677 | 67.8385 | 73.4375 | 74.349 |
| K-Star | 71.0938 | 65.625 | 70.4427 | 73.5677 |

Figure 1, depicts the performance of the discussed feature selection algorithms after classification for diabetes dataset. On the average EDR method exhibits highest classification accuracy and is the best unsupervised feature selection method for diabetes data set.

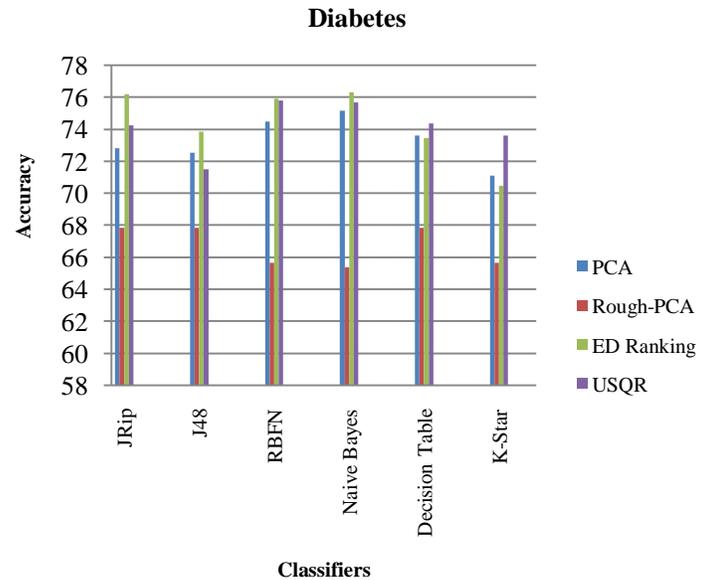

Figure 1: Classification Accuracy for Diabetes

TABLE 4: CLASSIFICATION ACCURACY FOR BREAST CANCER

| Classifiers | PCA | Rough-PCA | EDR | USQR |
|---|---|---|---|---|
| JRip | 91.5641 | 91.5641 | 79.4376 | 90.5097 |
| J48 | 91.2127 | 91.2127 | 78.2074 | 90.6854 |
| RBFN | 91.5641 | 91.5641 | 80.1406 | 90.3339 |
| Naive Bayes | 90.8612 | 90.8612 | 78.2074 | 89.8067 |
| Decision Table | 88.4007 | 88.4007 | 78.3831 | 90.6854 |
| K-Star | 91.0369 | 91.0369 | 77.3286 | 90.3339 |

TABLE 5: CLASSIFICATION ACCURACY FOR LUNG CANCER

| Classifiers | PCA | Rough-PCA | EDR | USQR |
|---|---|---|---|---|
| JRip | 87.5 | 65.625 | 81.25 | 62.5 |
| J48 | 87.8 | 59.375 | 81.25 | 56.25 |
| RBFN | 90.625 | 40.625 | 65.625 | 46.875 |
| Naive Bayes | 84.375 | 56.25 | 78.125 | 62.5 |
| Decision Table | 81.25 | 59.375 | 81.25 | 53.125 |
| K-Star | 90.625 | 50 | 71.875 | 56.25 |

Figure 2, depicts the performance of the discussed feature selection algorithms after classification for breast cancer dataset. On the average PCA and Rough-PCA method exhibits highest classification accuracy and is the best unsupervised feature selection method for breast cancer data set.

Figure 3, depicts the performance of the discussed feature selection algorithms after classification for lung cancer dataset. On the average PCA method exhibits highest classification accuracy and is the best unsupervised feature selection method for lung cancer data set.

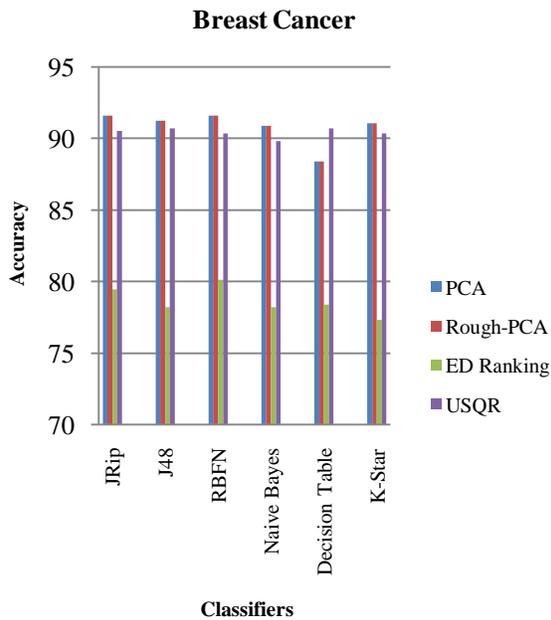

Figure 2: Classification Accuracy for Breast Cancer

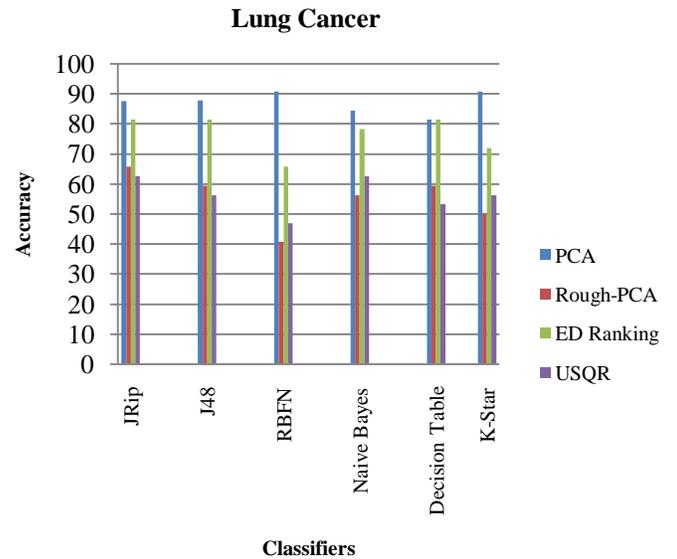

Figure 3: Classification Accuracy for Lung Cancer

TABLE 6: CLASSIFICATION ACCURACY FOR Ecoli

| Classifiers | PCA | Rough-PCA | EDR | USQR |
|---|---|---|---|---|
| JRip | 57.7381 | 58.0357 | 80.9524 | 78.2738 |
| J48 | 63.3929 | 63.0952 | 81.8452 | 77.381 |
| RBFN | 61.3095 | 58.631 | 81.25 | 78.381 |
| Naive Bayes | 61.3095 | 65.7738 | 85.4167 | 80.0595 |
| Decision Table | 62.5 | 62.2024 | 77.0833 | 76.4881 |
| K-Star | 66.369 | 56.4762 | 81.5476 | 79.7619 |

TABLE 7: CLASSIFICATION ACCURACY FOR HEART

| Classifiers | PCA | Rough-PCA | EDR | USQR |
|---|---|---|---|---|
| JRip | 73.3333 | 64.0741 | 70.0000 | 67.7778 |
| J48 | 75.1852 | 60.7407 | 66.6667 | 67.7778 |
| RBFN | 71.8519 | 65.5556 | 73.3333 | 69.2593 |
| Naive Bayes | 73.3333 | 67.037 | 71.4815 | 66.6667 |
| Decision Table | 71.4815 | 67.7778 | 70.3704 | 70.7407 |
| K-Star | 71.4815 | 63.3333 | 70.3704 | 64.0741 |

Figure 4, depicts the performance of the discussed feature selection algorithms after classification for Ecoli dataset. On the average EDR method exhibits highest classification accuracy and is the best unsupervised feature selection method for Ecoli data set.

Figure 5, depicts the performance of the discussed feature selection algorithms after classification for Heart dataset. On the average PCA method exhibits highest classification accuracy and is the best unsupervised feature selection method for heart data set.

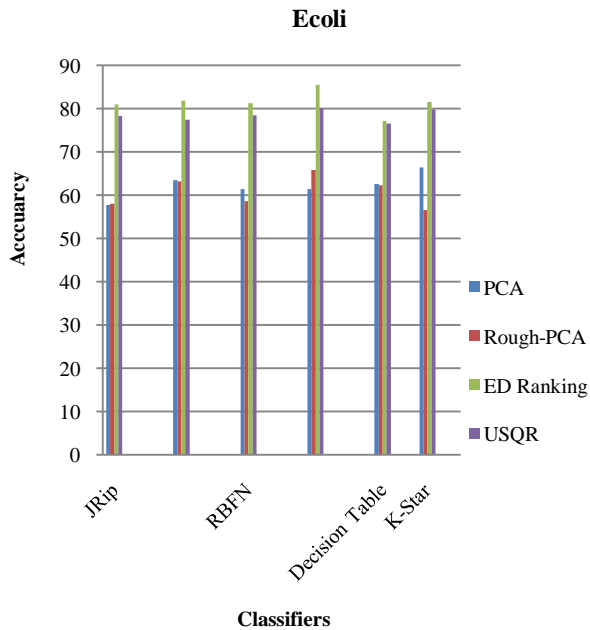

Figure 4: Classification Accuracy for Ecoli

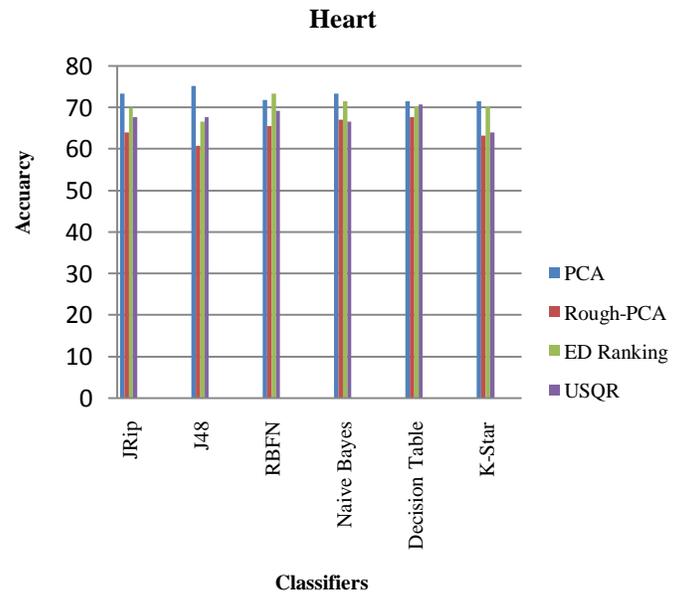

Figure 5: Classification Accuracy for Heart

## IV. CONCLUSION

In this paper, PCA, EDR, Unsupervised Quick Reduct and Rough-PCA based on rough set theory has been implemented on some synthetic and biological datasets from data repository. The WEKA tool is used to compute classification accuracy of the selected subset of features. EDR outperforms other methods for several data sets than other methods and has proven to be the best method for unsupervised feature selection.